\documentclass[runningheads]{llncs}
\usepackage[T1]{fontenc}
\usepackage{graphicx}
\usepackage{times}  % DO NOT CHANGE THIS
\usepackage{helvet}  % DO NOT CHANGE THIS
\usepackage{courier}  % DO NOT CHANGE THIS
\usepackage[hyphens]{url}  % DO NOT CHANGE THIS
\usepackage{graphicx} % DO NOT CHANGE THIS
\usepackage[numbers,sort&compress]{natbib}  % DO NOT CHANGE THIS AND DO NOT ADD ANY OPTIONS TO IT
\usepackage{caption} % DO NOT CHANGE THIS AND DO NOT ADD ANY OPTIONS TO IT
\usepackage{algorithm}
\usepackage{algorithmic}

\usepackage{newfloat}
\usepackage{listings}
\DeclareCaptionStyle{ruled}{labelfont=normalfont,labelsep=colon,strut=off} % DO NOT CHANGE THIS
\floatstyle{ruled}
\newfloat{listing}{tb}{lst}{}
\floatname{listing}{Listing}
\usepackage{bibentry}
\providecommand{\shortver}[1]{}
\shortver{\newcommand{\longver}[1]{}}
\providecommand{\longver}[1]{#1}

\usepackage{algorithm}
\usepackage{algorithmic}
\usepackage[utf8]{inputenc} % allow utf-8 input
\usepackage[T1]{fontenc}    % use 8-bit T1 fonts
\usepackage{xcolor}         % colors
\usepackage{url}            % simple URL typesetting
\usepackage{booktabs}       % professional-quality tables
\usepackage{amsfonts}       % blackboard math symbols
\usepackage{nicefrac}       % compact symbols for 1/2, etc.
\usepackage{microtype}      % microtypography
\usepackage{xcolor}         % colors
\usepackage{todonotes}%[disable]
\usepackage{relsize}
\usepackage{enumitem,kantlipsum}
\usepackage{multicol}

\usepackage{amsmath,amsfonts,bm}
\usepackage{subfigure}

\def\eqref#1{equation~\ref{#1}}
\def\1{\bm{1}}

\DeclareMathAlphabet{\mathsfit}{\encodingdefault}{\sfdefault}{m}{sl}
\SetMathAlphabet{\mathsfit}{bold}{\encodingdefault}{\sfdefault}{bx}{n}

\DeclareMathOperator*{\argmin}{arg\,min}

\usepackage{xspace}
\newcommand{\length}{\ell}
\newcommand{\tree}{T}
\newcommand{\policy}{\phi}
\newcommand{\satinstance}{S}

\newcommand{\kpath}{P}
\newcommand{\weight}{w}
\newcommand{\node}{n}
\newcommand{\nodes}{N}
\newcommand{\size}{s}
\newcommand{\depth}{d}

\newcommand{\variables}{X}
\newcommand{\variable}{v}
\newcommand{\clauses}{C}
\newcommand{\cpuct}{c_{PUCT}}
\newcommand{\lookaheads}{k}

\newcommand{\treepolicy}{\alpha}
\newcommand{\noaction}{\emptyset}
\newcommand{\rollout}{\theta}
\newcommand{\rolloutpolicy}{\pi}

\newcommand{\steppolicy}{\gamma}
\newcommand{\sibling}{\state^{sub}}
\newcommand{\subsolver}{\policy_{sub}}
\newcommand{\state}{s}
\newcommand{\action}{a}
\newcommand{\Ex}[2][]{\operatorname{\mathbb{E}}_{#1}{#2}}

\definecolor{1}{RGB}{231,71,121}
\definecolor{2}{RGB}{0,157,153}%{119,197,163}
\definecolor{3}{RGB}{65,115,182}
\definecolor{4}{RGB}{237,141,30}%{241,165,77}

\newcommand{\colourselection}{3}
\newcommand{\colourexpansion}{4}
\newcommand{\coloursimulation}{2}
\newcommand{\colourbackup}{1}

\shortver{
    \newenvironment{kequation}{$}{$}
    \newcommand{\minipar}[1]{\medskip\noindent\textbf{{#1.}}~}
    \newcommand{\superminipar}[1]{\smallskip\noindent\textbf{\emph{#1.}~~}}
    \newcommand{\micropar}[1]{\noindent\emph{#1.}~}
    \let\subsection\minipar
    \let\subsubsection\superminipar
    \let\paragraph\micropar
}
\longver{
    \newenvironment{kequation}{\begin{equation}}{\end{equation}}
}

\author{Chris Cameron\inst{1},
Jason Hartford\inst{2},
Taylor Lundy\inst{1},
Tuan Truong\inst{1},\\
Alan Milligan\inst{1},
Rex Chen\inst{3},
Kevin Leyton-Brown\inst{1}
}
\institute{\small Department of Computer Science, University of British Columbia, Vancouver, BC
\email{\{cchris13,tlundy,kevinlb\}@cs.ubc.ca, manhtuan15042000@gmail.com, alanmil@student.ubc.ca}
\and
 Valence Labs, Montr\'eal, QC
 \email{jason@valencelabs.com}
 \and
 School of Computer Science, Carnegie Mellon University, Pittsburgh, PA
 \email{rexc@cmu.edu}}

\newtheorem{thm}{Theorem}
\newtheorem{cor}{Corollary}

\authorrunning{C. Cameron \emph{et al.}}

\usepackage{relsize}  % Include the relsize package
\usepackage{letltxmacro}

\LetLtxMacro\oldtexttt\texttt
\renewcommand{\texttt}[1]{{\relsize{-0.5}{\oldtexttt{#1}}}}

\title{UNSAT Solver Synthesis \\via Monte Carlo Forest Search}

\begin{document}

\maketitle

\begin{abstract}
We introduce Monte Carlo Forest Search (MCFS), a class of reinforcement learning (RL) algorithms for learning policies in {tree MDPs}, for which policy execution  involves traversing an exponential-sized tree. Examples of such problems include proving unsatisfiability of a SAT formula; counting the number of solutions of a satisfiable SAT formula; and finding the optimal solution to a mixed-integer program. MCFS algorithms can be seen as extensions of Monte Carlo Tree Search (MCTS) to cases where, rather than finding a good path (solution) within a tree, the problem is to find a small tree within a forest of candidate trees. We instantiate and evaluate our ideas in an algorithm that we dub Knuth Synthesis, an MCFS algorithm that learns DPLL branching policies for solving the Boolean satisfiability (SAT) problem, with the objective of achieving good average-case performance on a given distribution of unsatisfiable problem instances. Knuth Synthesis is the first RL approach to avoid the prohibitive costs of policy evaluations in an exponentially-sized tree, leveraging two key ideas: first, we estimate tree size by randomly sampling paths and measuring their lengths, drawing on an unbiased approximation due to Knuth (1975); second, we query a strong solver at a user-defined depth rather than learning a policy across the whole tree, to focus our policy search on early decisions that offer the greatest potential for reducing tree size. We matched or exceeded the performance of a strong baseline on three well-known SAT distributions, facing problems that were two orders of magnitude more challenging than those addressed in previous RL studies.

\end{abstract}

\section{Introduction}
\label{sec:introduction}

\citet{Silver2017,Silver2018} took the world by storm when their AlphaGo system beat world champion Lee Sedol at Go, marking the first time a computer program had achieved superhuman performance on a game with such a large action space. Their key breakthrough was combining Monte Carlo Tree Search (MCTS) rollouts with a neural network-based policy to find increasingly strong paths through the game tree. This breakthrough demonstrated that, with good state-dependent policies, MCTS can asymmetrically explore a game tree to focus on high-reward regions despite massive state spaces. 
MCTS rollouts avoid the exponential cost of enumerating all subsequent sequences of actions \citep{Kearns2002,Coulom2006} and can be extremely efficient when leveraging (1) multi-arm bandit policies  to trade off exploration and exploitation \citep{Kocsis2006} and (2) function approximation of the policies and values from previous problems to provide priors that further focus rollouts on promising paths \citep{Coulom2007,Sutskever2008,Maddison2014}.

\begin{figure}[t]
    \centering
    \includegraphics[width=\columnwidth, trim={0 11pt 0 2pt},clip]{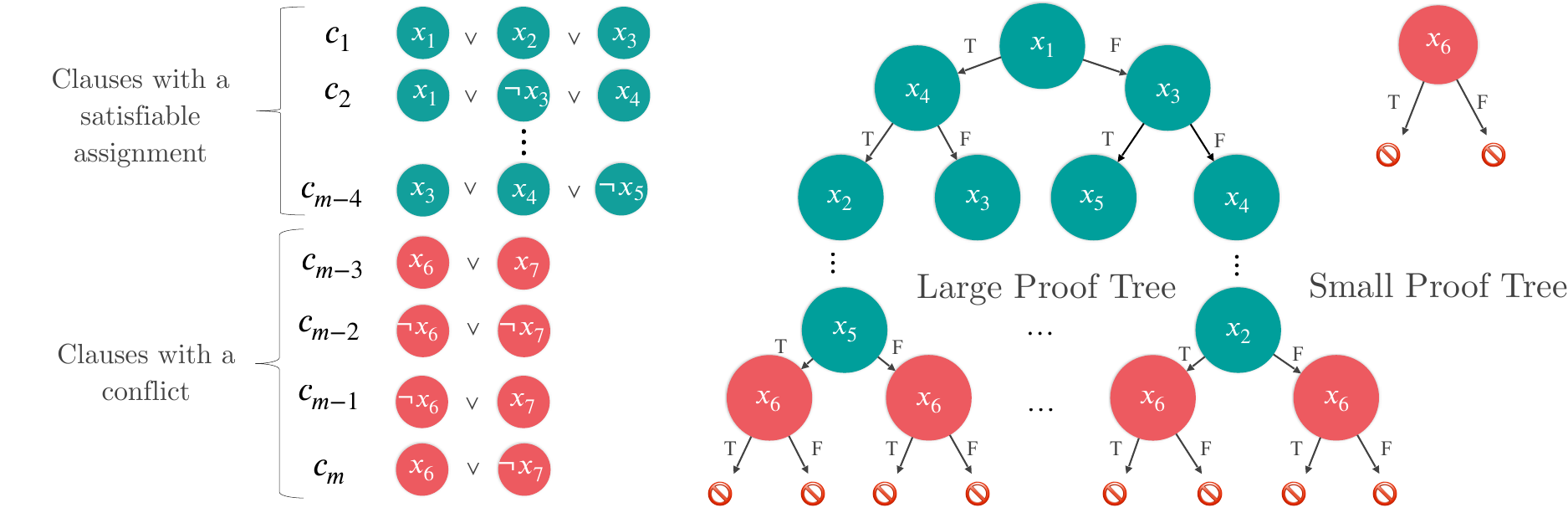}
    \caption{A proof tree shows that any variable assignment to an unsatisfiable instance leads to a conflict. Its size substantially varies with the choice of branching policy.}
    \label{fig:var_branching}
\end{figure}

MCTS is most useful in combinatorial spaces that have a natural hierarchical decomposition into a search tree. As a result, most of the notable applications of MCTS were to search for good paths of actions through game trees \citep{Coulom2006,Baudis2012,Zook2019,Agostinelli2019,Schrittwieser2020}. MCTS is also useful for searching for solutions to NP-hard combinatorial problems where a solution is a path through a search tree \citep{Previti2011,Browne2012,Abe2019,Khalil2022}. 
However, many other combinatorial problems cannot be expressed as searching for a good path. Consider constraint satisfaction problems (CSPs): while solving a satisfiable problem corresponds to finding a path (assigning variables sequentially and checking the solution), existing methods for proving that no solution exists build out a \emph{proof tree} demonstrating that all possible variable assignments lead to a conflict. Rather than finding a high-reward path within a single tree, an algorithm designer's goal is to find a small tree within the forest of possible trees (e.g., see Figure \ref{fig:var_branching}). % defined by the space of branching policies. 
This difference matters because of the high cost of evaluating each candidate policy: individual trees can be exponentially large. 
 
This paper introduces Monte Carlo Forest Search (MCFS), a class of MCTS-inspired RL algorithms for such settings. We leverage \citet{Scavuzzo2022}'s recent concept of \emph{tree MDPs}. When an action is taken in a tree Markov Decision Process (MDP), the environment transitions to two or more new child states; each of these new child states satisfies the Markov property in that each child only depends on its state. Algorithms that recursively decompose problems into two or more simpler subproblems in a history-independent way can be represented as tree MDP policies. 
An MCFS rollout is a policy evaluation in the tree MDP and hence builds out a tree rather than a path, with the value of each node corresponding to the sum of rewards over its descendants in the rollout tree. Similar to MCTS, MCFS is a broad class of algorithms that can be adapted for different domains and problem sizes through choices in how policies are evaluated, how actions are selected and how rewards are aggregated. For example, an MCFS algorithm with complete policy evaluation (evaluating every child in a rollout) would be sufficient for tree MDPs giving rise to extremely small policy trees (e.g., log-linear trees in MergeSort). However, we are interested in tree MDP problems where policies produce exponential-sized trees, making exact policy evaluation prohibitive. We are not aware of any method that addresses this issue; existing work using RL \cite{Kurin2020,Lederman2019,Vaezipoor2021,Scavuzzo2022,parsonson2022reinforcement,etheve2020reinforcement} require exact policy evaluation and have only been used to train on very easy instances by industry standards (solvable on the order of 1000 decisions). 

\begin{figure}[t]
    \centering
    \includegraphics[width=\columnwidth, trim={0 11pt 0 1pt},clip]{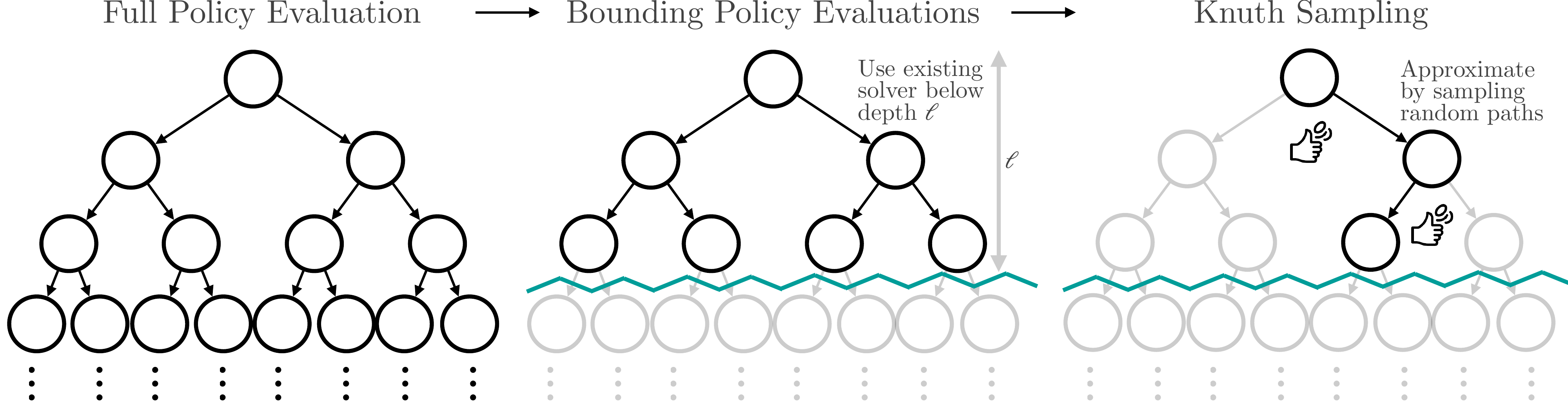}
    \caption{Two keys ideas of Knuth Synthesis to avoid the prohibitive costs of exact tree-size policy evaluations: (1) bounded-depth search and (2) Knuth samples.}
    \label{fig:knuth_synthesis}
\end{figure}

To solve this problem, we present Knuth Synthesis, an MCFS algorithm that allows for cheaper policy evaluation and is tailored for  \emph{pure-DPLL} \citep{davis1962machine} algorithms, a popular class of SAT algorithms that can straightforwardly be defined as tree MDPs.\footnote{Other more practically useful algorithms such as CDCL and Branch-and-Bound cannot easily be represented as tree MDPs because they correspond to deeply history-dependent policies; we are not aware of any RL methods that are tractable for learning policies for such algorithms.} We use the idea of \emph{Knuth sampling} \citep{Knuth1975} to obtain linear and unbiased Monte Carlo approximations of tree sizes, following \citet{Lobjois1998} who showed these can be effective for cheaply comparing algorithms despite their high variance. Knuth samples correspond to path-based rollouts, allowing us to integrate them with MCTS. Extensive follow-up work has developed alternatives for approximating tree size (e.g., \cite{Purdom1978,Chen1992,Cornuejols2006,Kilby2006}), but these estimates are not decomposable into path-based root-to-leaf rollouts, and would require new ideas to be integrated into MCFS. Knuth Synthesis is designed to be used offline to synthesize new solvers under the data-driven algorithm design paradigm \cite{DBLP:journals/corr/abs-2011-07177}, where we optimize algorithm performance over a training set of instances.
As is standard in the literature, we encode our policies using deep neural networks, which are far more expensive to evaluate than standard heuristics. Such policies are too expensive to evaluate at every node of the search tree. We mitigate this cost by limiting our learned policy to the search tree's most important nodes, querying some existing subsolver below a certain depth in the proof tree; this also substantially reduces the policy space. To ensure that we find strong policies for online use, we enforce the same procedure offline for constraining on which nodes to call the policy. Figure \ref{fig:knuth_synthesis} illustrates the two keys ideas for making policy evaluation tractable. % 
  
Knuth Synthesis is the first MCTS-like method that has been used to learn a branching policy for combinatorial search. 
We deviated from standard MCTS ideas as appropriate, notably making a novel change to the bandit algorithm to account for our tree size cost function. Unlike AlphaZero, which uses MCTS both offline and online, we use Knuth Synthesis only offline, as we cannot afford the computational overhead online; we only query our policy online.  

We evaluated our method on specific unsatisfiable problem distributions where $\approx$ 100,000 decisions were required to solve problems during training, a considerable leap beyond other prominent published work. We note that focusing on either satisfiable or unsatisfiable instances in commonplace in SAT. There are many examples of specialized algorithms targeting one of satisfaible or unsatisfiable instances; indeed, there have been specialized tracks of the SAT competition for both cases. For example, local search is a very well-studied class of algorithms that cannot prove unsatisfiability. In practice, one can always use a portfolio strategy running specialized SAT and UNSAT algorithms in parallel and terminating when either returns an answer. Solving such problems efficiently is important in practice, e.g., for system debugging \citep{Suelflow2008} and formal verification \citep{Bryant2009}.

First, we evaluated our method on uniform random 3-SAT at the solubility phase transition, perhaps the best-studied SAT distribution (e.g., featured in the SAT Competition \citep{SATcomp} from 2002 to 2018). We matched the performance of \texttt{kcnfs}, which was specifically designed to target this distribution. Second, we evaluated our method on the \texttt{sgen} \citep{Spence2010} distribution, which is notoriously difficult for its problem size. We improved running time on \texttt{sgen} by 8\% over \texttt{kcnfs}, which in turn is $3.2\times$ faster on this distribution than the \texttt{hKis} solver that solved the most unsatisfiable instances in the 2021 SAT Competition. Lastly, we evaluated our method on the \texttt{satfc} \citep{Frechette2016} distribution, consisting of radio-spectrum repacking feasibility problems from an FCC spectrum auction, improving running time by 28\% over \texttt{kcnfs}. These results show the initial promise of MCFS; through further scaling, we believe these ideas will lead to stronger industrial solvers for specific applications. 

\section{Related Work} 
\label{sec:related-work}
\subsubsection*{Branching}Tree search is a fundamental algorithm for combinatorial optimization that iteratively partitions a search space, e.g., by choosing variables to branch on. At a high level, the efficiency of tree search is measured by the product of (1) the number of branches and (2) the time required to make each branching decision. The order in which variables are assigned—the branching policy—has a dramatic effect on the size of the tree and the corresponding time to solve the problem, so algorithm designers seek branching policies that lead to small trees. In domains such as Mixed Integer Programming (MIP) solving, a ``smart and expensive'' paradigm has won out, where entire linear programs are solved to determine a branching decision. In other domains such as SAT solving, a ``simple and cheap'' paradigm dominates: high-performance SAT solvers make very cheap branching decisions that depend only superficially on subproblem state. After 20 years, Variable State Independent Decaying Sum (VSIDS) \citep{Moskewicz2001} has remained among the state of the art %(how often variable occurs in conflict analysis (recency weighted). 
for Quantified Boolean Formula (QBF) and SAT; it is a simple heuristic that only keeps track of how often a variable occurs in conflict analysis.
Some interesting early work developed more expensive heuristics \cite{Huang2003}, but these generally could not compete with VSIDS in terms of running time. %creates DTrees from CNF that maximizes disjointness over different parts of tree. Roughly want to choose variables that makes two subproblems most disjoint from each other.
A few notable exceptions are: (1) look-ahead SAT solvers, which have a similar concept of a lookahead as MCFS and choose a branching variable after looking ahead at the resulting state for different possible branches \cite{heule2009look, heule2011cube}; (2) the bsh heuristic used in the \texttt{kcnfs} solver \citep{Dequen2003} which branches based on a measure of how constraining the assignment of one variable is on the assignment of another; (3) Bayesian moment matching to initialize VSIDS per-variable scores for literals based on how likely they are to be part of a satisfying assignment \cite{Duan2020}.

There is now a long history of machine learning being used for automated algorithm design \cite{hoos2012automated,kotthoff2016algorithm,cappart2023combinatorial, bengio2021machine}.
One might expect that machine learning could learn more informative branching heuristics for tree search that are worth their cost, especially for shallow-depth branches where branching decisions are most consequential. %\cc{Need some text about why shallow-depth decisions more important} 
Two approaches have shown promise for learning models to make branching decisions: imitation learning and reinforcement learning. 

First, imitation learning works by learning a cheap approximation of (1) an expensive existing heuristic \cite{khalil2016learning,Gasse2019,hottung2020deep, Nair2020} or (2) an expensive feature that is a good proxy for a good branching decision \cite{wang2022neurocomb, Selsam2019}. \citet{khalil2016learning} learned a cheaper approximation of the strong branching heuristic in MIP using hand-crafted variable-level features to achieve state-of-the-art performance; \citet{Gasse2019} and \citet{Nair2020} learned an improved approximation of strong branching by learning a feature representation by representing MIPs as graphs and leveraging graph neural networks.  
Relevant to SAT, \citet{wang2022neurocomb} predict backdoor variables and integrate these predictions into the VSIDS heuristic while \citet{Selsam2019} learned to approximate small unsatisfiable core computation and then branched on variables predicted to belong to a core. Both report practical performance improvements. \citeauthor{Selsam2019}'s work is particularly relevant to the UNSAT setting, which we focus on. They computed small unsatisfiable cores for 150,000 instances and trained a neural network to predict them, achieving a 6\% speedup over a strong SAT solver when using their network for a limited number of top-level branching decisions. Note that if there exists a small unsatisfiable core, there exists a corresponding short proof, but not vice versa. In other related SAT work, \citet{Nejati2020} trained a branching policy to predict the branching variable that leads to the smallest running time from a single branch starting at the root of the tree with a fixed downstream solver. 
Their motivating application is parallel SAT solving where each branch of the tree are solved in parallel.

Second, it is tempting to use reinforcement learning to directly synthesize heuristic policies that are optimized for problem distributions of interest. This idea dates back to \citet{Lagoudakis2001}, who used TD-learning to train a policy to select between seven predefined branch heuristics based on simple hand-crafted features. With the advent of modern deep learning architectures, practitioners train models that take the raw representation of a problem instance as the model input.  \citet{Yolcu2019} learned a local search heuristic for SAT; \citet{Tonshoff2022} built a generic graph neural network-based method for iteratively changing assignments in CSPs;
\citet{Lederman2019} used the REINFORCE algorithm to learn an alternative to the VSIDS heuristic for QBF that improved CPU time within a competitive solver; \cite{song2022learning} used a Q-learning approach for to learn a branching policy for CSP;
and \citet{Vaezipoor2021} used a black-box evolutionary strategy to learn a state-of-the-art branching heuristic for model counting, demonstrating improvements in walltime performance over the competitive \texttt{SharpSAT} solver. 
There are also various applications of RL for learning to branch in MIP 
\cite{Scavuzzo2022, etheve2020reinforcement, parsonson2022reinforcement}.
Finally, \citet{Kurin2020}'s work, which used Q-learning for branching in the CDCL solver Minisat, is closest to our own. They trained on random satisfiable problems with 50 variables and generalized online to 250 variables for both satisfiable and unsatisfiable instances and improved Minisat running time. Instances in their training set required on the order of 100 decisions. The authors found that increasing the difficulty of the training set generalized poorly and training was less stable; the authors cite higher variance returns from longer episodes, challenges for temporal credit assignment, and difficulties with exploration. They resolved the path/tree distinction by treating a traversal through a tree as a path (which could be exponential in length) and allowing backtracking state transitions. This, coupled with not incorporating the stack of backtracking points into their state violates the Markov assumption when backtracking, which is necessary for proving unsatisfiability and is important for satisfiability for all but the optimal policy.
%Their value function needs information about total tree, but they only condition on variables and clauses at sub state. 
%\cc{TODO: It's not just that traversals are of exponential length. Given their state definition, they don't have an MDP, because a state can transition (backtrack) to any arbitrary other state given the larger problem that the subproblem lies within. I find this hard to say succinctly} 

We think that the problem of using RL to train a branching policy deserves study even if the field remains a few steps behind practical relevance. The primary challenge arising in this body of past work is scaling RL methods to more practical problems requiring more decisions. Each of these RL methods do exact policy evaluation and have only been trained with $\leq 1000$ decisions and are therefore constrained to training on easy instances. From the perspective of a modern SAT solver, many of these results may appear trivial, but they nevertheless quantify the extent to which RL methods can learn policies that improve on hand-tuned heuristics.

\subsubsection*{RL with trees}\citet{Scavuzzo2022} were the first to cleanly pinpoint how the structure of the tree-search algorithm changes credit assignment in RL, introducing the concept of \emph{tree MDPs} and formulating a policy gradient for tree policies. Much earlier, \citet{Lagoudakis2001} recognized that the one-to-two state transition violated the MDP definition and resolved this by cloning the MDP and creating one copy for each transition. 
A number of other recent papers have also realized the inefficiency of credit assignment when treating an episode as a path (rather than a tree) in a tree-search algorithm \citep{etheve2020reinforcement, parsonson2022reinforcement, song2022learning}.

\subsubsection*{Approximating Tree Size}We approximate the tree size of a DPLL policy via \emph{Knuth samples} \citep{Knuth1975}, which \citet{Lobjois1998} showed can be effective for cheaply comparing algorithms despite its high variance. A Knuth sample provides a path-based rollout, which allows us to easily leverage MCTS. There has been extensive follow-up work developing alternatives for approximating tree size (e.g., \citet{Purdom1978,Chen1992,Cornuejols2006,Kilby2006}), but these estimates are not decomposed into path-based rollouts from root to leaf, and require new ideas to be integrated into MCFS.

\subsubsection*{MCTS for Combinatorial Problems}Various researchers have used MCTS as an algorithm to directly search for a satisfying solution to a CSP that lie somewhere between local search and DPLL \cite{Previti2011,Schloeter2017,Keszocze2020,Loth2013}. In these cases, a rollout can be interpreted as a guess at a satisfying assignment. If an unseen node is reached or a conflict is reached along a rollout path, a reward is assigned based on some measure of how close the path is to being a satisfying assignment (e.g., number of satisfied constraints). We see two main differences between these approaches and ours: (1) they use MCTS online to solve CSPs rather than as an offline procedure for training model-based branching policies and (2) they are not designed for the unsatisfiable case where policies produce trees rather than paths.

\citet{Previti2011} developed UCTSAT, which assigns 0 reward to a conflict node and explores various alternatives for value estimation at non-terminal nodes such as (1) number of satisfied clauses at current state and (2) average number of satisfied clauses based on random paths from that state. A number of follow-up works have incorporated clause learning into UCTSAT. \citet{Schloeter2017} added new clauses to the problem for every conflict that is encountered. This addition does not affect the UCT tree since state is defined as the set of assigned variables and therefore independent of additional clauses. \citet{Keszocze2020} made better use of the learned clauses, filtering out any expanded nodes that are ruled out by a learned clause. They modified \citeauthor{Previti2011}'s reward function by weighting each satisfied clause by its activity (number of times occurring in resolution) and propose a number of other variations including penalizing by conflict depth; we penalized by $2^{\text{depth}}$, which corresponds to minimizing the size of the proof tree. \cite{Loth2013} developed a version of UCTSAT for Constraint Programming (CP) that was simplified to be state independent. Rather than each tree node representing a different multi-armed bandit (MAB), they have a single MAB that they update at every node of the tree. This change was made to incorporate restarts, where there would be too few samples for each node. They reward a path by the length until conflict. \citet{Wattez2020} used a single MAB to choose which branching heuristic to use on a given instance. They rewarded a path to a conflict by the number of assignments ruled out by that path. 

Outside of CSP, there are a number of other existing uses of MCTS for solving \emph{path-based} NP-hard problems. \citet{Browne2012} used the UCT algorithm for solving MIPs, taking paths from the root to a leaf and propagating up the maximum over child LP values; \citet{Abe2019} searched over assignment paths in graph problems such as choosing edges to cut in a graph; and \citet{Khalil2022} searched over paths of variables to find MIP backdoors. 

\subsubsection*{Relation to conventional ML for SAT}There is a significant amount of work over the past few decades in ML for SAT using algorithm selection \cite{kotthoff2016algorithm} and algorithm configuration \cite{hoos2012automated}, however we do not leverage those ideas in our work. We think that RL for learning a branching policy is conceptually distinct from selection and configuration. We see selection and configuration as hand-engineered ML; they rely on either experts exposing algorithm parameters or instance features. On a spectrum of ``degree of automation”, we think ``RL for branching” falls somewhere between hand-engineered ML and pure end-to-end approaches (e.g., NeuroSAT \cite{Selsam2019}) where the network directly outputs a SAT solution. We leverage the scaffolding of existing tree-search solvers but we are flexible to learn any arbitrary state-dependent policy within that space. We could in principle define our problem of learning a branching heuristic as an algorithm configuration problem (i.e., the neural net parameters are our algorithm parameters) but most algorithm configuration approaches use black-box optimization whereas we are opening up the black box by setting it up as an RL problem.

We also note that our method is compatible with algorithm selection or configuration at the level of the \emph{subsolver}. Knuth Synthesis is agnostic to the choice of subsolver; we could in principle train an algorithm selector or configure a solver as the subsolver and we think there could be some exciting synergies there. For example, we considered a novel algorithm selection setting where we would select amongst solvers for different subproblems within a single instance, rather than the typical setting where one algorithm is selected per instance. That would elicit a new design dimension; we would want to find branching policies that produce subproblems that can exploit the complementary of an algorithm portfolio.

\section{Preliminaries} 
\label{sec:preliminaries}
We now provide the required technical background for MCTS, the Boolean satisfiability problem (our application area), the DPLL algorithm (the framework that defines our policy space), and tree MDPs (the class of problems to which we apply MCFS).

\subsection{Monte Carlo Tree Search}
\label{sec:preliminaries:mcts}
Monte Carlo Tree Search (MCTS)
 is a general-purpose RL algorithm framework for MDPs. An MDP $M(\mathcal{S}, \mathcal{A}, p, r)$ is defined by a set of states $s \in \mathcal{S}$, actions $a \in \mathcal{A}$, transition distribution $p(s_{i+1}| s_{i}, a_{i})$, and reward function $r : \mathcal{S}\times\mathcal{A} \rightarrow \mathbb{R}$. For every state $s_{i}$ that is visited, MCTS stores a vector of counts $c_{i}\in \mathbb{Z}^{|\mathcal{A}|}$ and value estimates $v_{i}\in \mathbb{R}^{|\mathcal{A}|}$. A \emph{rollout} $\rollout$ is a sequence of state, action pairs $((s_{0},a_{0}),...,(s_{n}, a_{n}))$. MCTS makes a series of rollouts starting from the current state $s_{0}$ defined with the below four steps. Our definition is more general than presented in \citet{sutton2018reinforcement} to help us later define MCFS, notably the introduction of $\steppolicy$ that separates the concepts of (1) choosing an action from (2) terminating a rollout.

\begin{enumerate}%[wide, labelwidth=!, labelindent=0pt]
    \item {\bf Selection.} A \emph{tree policy} $\treepolicy:s\in \mathcal{S},c \in \mathbb{R}^{|\mathcal{A}|}$,$v \in \mathbb{R}^{|\mathcal{A}|}\mapsto\mathcal{A}$ selects an action based on the action values $v$, counts $c$, and often a prior that depends on $s$. Then a \emph{step policy} $\steppolicy:\rollout \in (\mathcal{S}\times\mathcal{A})^{n}  ,\action \in \mathcal{A},\state \in \mathcal{S}\mapsto \mathcal{A}\cup \emptyset$ takes in the state $s_{i}$, action $a_{i}$, and the current rollout history $((s_{0},a_{0}),...,(s_{i}, a_{i}))$ and chooses to either play the action selected by $\treepolicy$ or to play no action ($\emptyset)$, terminating the rollout at that state. %Unlike $\treepolicy$, $\steppolicy$ is allowed to depend on the history of the rollout. 
    
    \item {\bf Expansion.} Counts and value estimates are tracked for any state reached by our selection step and together form a tree rooted at the current state. For any previously unexplored state $s_{i}$ along the path, we add a node with vectors $c_{i}$ and $v_{i}$ to our tree. 
    \item {\bf Simulation.} \emph{rollout policy} $\rolloutpolicy:\mathcal{S}\mapsto \mathbb{R}$ then estimates the sum of rewards before reaching terminal state of the MDP from the node where no action was played . $\rolloutpolicy$ is commonly a value network estimating the value of a path originating from the node's state. 
    \item {\bf Backup.} Pass back rewards through the path of the MCTS tree, so that $v_{i}$ at $s_{i}$ are updated with the sum of rewards from its descendants: $\rolloutpolicy(s_{n}) + \Sigma_{x=i+1}^{n-1} r(s_{x})$.
\end{enumerate}
After a fixed number of rollouts, MCTS commits to the action at the root according to accumulated statistics of the tree (e.g., for AlphaZero, the largest number of samples) and the resulting state becomes the new root node. This procedure repeats until MCTS finds a path to a leaf of the tree. It is then possible to train a policy network to approximate MCTS by using the counts $c_{i}$ and values $v_{i}$ of each node along this path as training examples. The policy network can then be used within $\treepolicy$ to further focus rollouts on promising paths as in AlphaZero.

\subsection{Boolean Satisfiability}
\label{sec:preliminaries:sat}
A Boolean satisfiability (SAT) instance $\satinstance$ is defined by a set of clauses $\clauses = \{c_1, \dots, c_m\}$ over a set of variables $\variables = \{x_1, \dots, x_n\}$. Each clause consists of a set of Boolean \emph{literals}, defined as either a variable $x_i$ or its negation $\neg{x_i}$. Each clause is evaluated as \emph{True} iff at least one of its literals is true (i.e., the literals in a clause are joined by OR operators). For example, a clause $c_i = x_j \vee \neg x_k$ evaluates to \emph{True} iff either $x_j$ is set to \emph{True} or $x_k$ is set to \emph{False}. $\satinstance$ is \emph{True} if there exists an assignment of values to variables for which all the clauses simultaneously evaluate to \emph{True} (i.e., the clauses are joined by AND operators). If such an assignment exists, the instance is called \emph{satisfiable}; it is called \emph{unsatisfiable} otherwise. SAT solvers try to find an assignment of variables to demonstrate that a problem is satisfiable, or to construct proofs showing that no setting of the variables can satisfy the problem. We consider only unsatisfiable problems. 

\subsection{DPLL and Variable Selection Policies}
\label{sec:preliminaries:dpll}
Many SAT solvers rely on the Davis-Putnam-Logemann-Loveland algorithm (DPLL), which assigns variables in an order given by some (potentially state-dependent) variable selection policy. 
\begin{definition}
Let $\satinstance$ be a SAT instance or any subproblem within a larger instance. A policy $\policy$ is a mapping $\policy:\satinstance \rightarrow (\variable)$ that determines which variable $\variable$ to assign in DPLL.
\end{definition}
Given a policy, the DPLL algorithm selects a variable to branch on and recursively checks both the \emph{True} and \emph{False} assignments in a tree-like fashion, performing \emph{unit propagation} at each step. Unit propagation assigns variables forced by single-variable (\emph{unit}) clauses; propagates them to other clauses; and repeats until no unit clauses remain. Each recursive DPLL call terminates when a \emph{conflict} (variable assignments form a contradiction) is found, forming a \emph{proof tree} (e.g., Figure \ref{fig:var_branching}). 
There can be massive gaps in performance between variable selection policies. For example, Figure \ref{fig:var_branching} shows a formula that leads to a three-node proof tree if $x_6$ is selected first by the policy (assigning $x_6$ results in $x_7 \wedge \neg x_7$, implying a contradiction), but a tree that could have as many as $2^{|\variables|} - 1$ nodes if $x_6$ is selected last. This also illustrates how top-level decisions are much more powerful; the proof tree doubles in size for every level at which a good branching decision (branch on $x_{6}$) is not made. Another way that policies can affect tree size is through DPLL's unit propagation step: policies that cause more unit propagation earlier in the search require fewer decisions overall and therefore yield smaller search trees.

For an instance $\satinstance$ solved using variable selection policy $\policy$, we denote the size of the resulting proof tree as $\tree_{\policy}(\satinstance)$. For a given distribution over problems $\mathbb{P}$, our goal is to find a policy $\policy^*$ that minimizes the average proof tree size 
$\mathcal{L}(\policy; \mathbb{P}) = \Ex[\satinstance'\sim \mathbb{P}][\tree_{\policy}(\satinstance')]$.
Finding policies is computationally challenging; for an $n$-variable problem, there are $O(n^{3^n})$ possible variable selection policies ($3^{n}$ states representing each variable as \emph{True}, \emph{False}, or unassigned, and $n$ choices per state); and exact evaluation of $\tree_{\policy}(\satinstance)$ takes $O(2^{n})$ operations; \citet{liberatore2000complexity} showed that identifying the optimal DPLL branching variable at the root decision is both NP-hard and coNP-hard.\footnote{\citet{liberatore2006complexity} later showed the problem is $\Delta_{2}^{p}[log(n)]$-hard i.e., at least poly time with $log(n)$ oracle queries to an NP problem.} 
If we assume that the optimal variable selection policy, $\policy^*_\mathbb{P} = \argmin_{\policy'} \mathcal{L}(\policy'; \mathbb{P})$, is learnable by an appropriate model family, we could in principle learn an approximation to the optimal policy, $\smash{\hat{\policy}^*}$, to use within our solver. The challenge is to design a procedure that efficiently minimizes $\mathcal{L}(\policy; \mathbb{P})$ so that labeled training examples can be collected through rollouts of learned policies. This is a challenging reinforcement learning problem. For a particular variable selection policy, even evaluating the loss function on a single instance takes time exponential in $n$, and the number of policies is doubly exponential in $n$. 

\subsection{Tree MDPs} Tree MDPs \citep{Scavuzzo2022} are a generalization of MDPs with 1-to-many transitions. Similar to the Markov property of MDPs, tree MDPs have a \emph{tree Markov property}: each subtree depends only on its preceding state and action. We show how the DPLL algorithm can be represented with tree MDPs in Appendix \ref{app:dpll}.
\begin{definition}[\citet{Scavuzzo2022}]
\label{def:treemdp}
Tree MDPs are augmented MDPs $tM =(\mathcal{S}, \mathcal{A},$ $p_{init}, p^{L}_{ch}, p^{R}_{ch}, r, l)$,
with states $s \in \mathcal{S}$, actions $a \in \mathcal{A}$, initial state distribution $p_{init}(s_{0})$, left and right child
transition distributions $p^{L}_{ch}(s_{ch_{i}}^{L}|s_{i}, a_{i})$ and $p^{R}_{ch}(s_{ch_{i}}^{R}|s_{i}, a_{i})$, reward function $r : \mathcal{S}\times\mathcal{A} \rightarrow \mathbb{R}$ and leaf
indicator $l : \mathcal{S} \rightarrow \{0, 1\}$. Each non-leaf state $s_{i}$ (i.e., such that $l(s_i) = 0$), together with an action $a_i$, produces two new states $s_{ch_{i}}^{L}$ (left child) and $s_{ch_{i}}^{L}$ (right child). Leaf states (i.e., such that $l(s_i) = 1)$ are the leaf nodes of the tree, below which no action can be taken.
\end{definition}

\section{Monte Carlo Forest Search}
\label{sec:mcfs}

\begin{figure}[H]
    \includegraphics[width=\columnwidth, trim={0 80pt 0 60pt},clip]{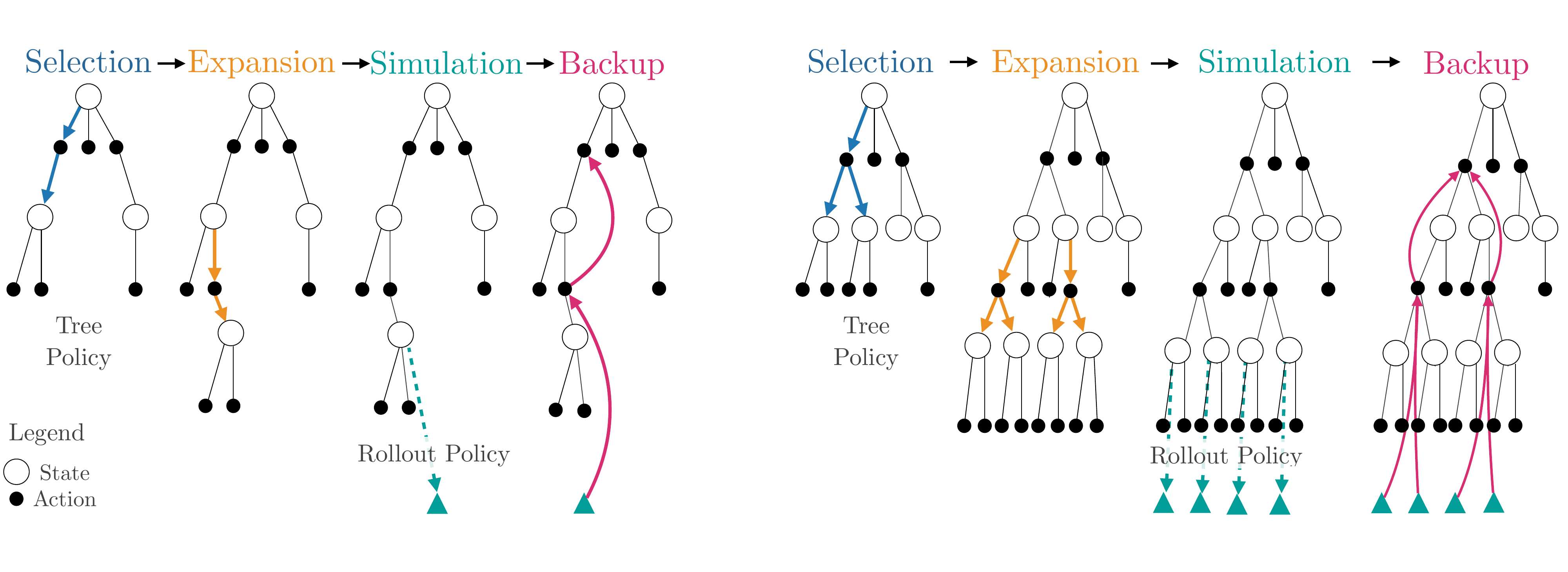}
    \input{mcfs_pseudocode}
    \caption{Side-by-side comparison of MCTS and MCFS highlighted by the \textcolor{\colourselection}{Selection}, \textcolor{\colourexpansion}{Expansion}, \textcolor{\coloursimulation}{Simulation} and \textcolor{\colourbackup}{Backup} steps (Adapted from \citet{sutton2018reinforcement}).} 
    \label{fig:mcfs}
\end{figure}

We define Monte Carlo Forest Search (MCFS) as a class of RL algorithms for learning policies in \emph{tree MDPs}. An MCFS rollout $\theta$ is a tree of (state, action) pairs structured according to the underlying tree MDP where a parent node $s_{i}$ shares edges with its children  $s_{ch_{i}}^{L}$ and $\smash{s_{ch_{i}}^{R}}$.  The aggregation of  \emph{rollout trees} forms a forest, hence the name.  The Expansion step is defined the same as MCTS and so is the Simulation step with the exception of being applied to many states rather than just one. We define the Selection and Backup step for an MCFS rollout below. See Figure \ref{fig:mcfs} for a side-by-side illustration ans pseudocode  of MCTS and MCFS in their most general forms.

{\bf Selection.} When an action selected by $\treepolicy$ is played by $\steppolicy$ at state $s_i$, the state transitions to new states $s_{ch_{i}}^{L}$ and $s_{ch_{i}}^{R}$ from sampling $p^{L}_{ch}(s_{ch_{i}}^{L}|s_{i}, a_{i})$ and $p^{R}_{ch}(s_{ch_{i}}^{R}|s_{i}, a_{i})$. These new states are added to a queue $q$. MCFS iterates over $q$, calling $\treepolicy$ and adding the corresponding new states back to $q$ if $\steppolicy$ chooses to step and terminates when $q$ is empty. 
 
{\bf Backup.} The rewards are passed back through $\rollout$, so that the value estimate $v_{i}$ at $s_{i}$ is updated with the sum of rewards from its descendants in $\rollout$.

We can now see the importance of defining the step policy $\steppolicy$; it allows us to evaluate any subset of the full proof tree in a \emph{rollout tree} which is important when exact policy evaluation is intractable. %We use Knuth sampling in or method that follows.  
MCFS commits to the action $a$ at the root in the same way as MCTS does and the corresponding  child nodes $s_{ch_{0}}^{L}$, $s_{ch_{0}}^{R}$ become the roots of new search trees and are solved sequentially. This procedure repeats until MCFS finds a tree where every leaf $s_i$ of the tree is a tree MDP leaf ($l(s_i)=1$).

\section{Knuth Synthesis}
\label{sec:methods}

For tree MDP problems like DPLL for UNSAT or model counting,  the tree depth can be linear in the number of actions; therefore the tree size (and thus the cost of evaluating a policy) can grow exponentially with the number of actions. Knuth Synthesis is an MCFS implementation based on two key ideas that avoid the prohibitive costs of exact policy evaluations (See Figure \ref{fig:knuth_synthesis}). First, in the step policy we nest a sampling procedure within a given Monte Carlo rollout, using a set of random paths through the rollout tree to approximate tree size. Second, we bound the depth of our forest search; below this point, we call out to a rollout policy that leverages existing SAT solvers. We describe these ideas in the following subsections and then provide our full implementation.

\subsection{Nested Monte Carlo Sampling}
To obtain a Monte Carlo sample through our tree MDP and determine the associated tree size $\tree_{\alpha}(\satinstance)$, we may need to visit $O(2^{n})$ states.
This is computationally expensive, both because of the absolute number of states and because at each node in the tree we need to query $\treepolicy$ to decide which variable to assign next (which is nontrivial when $\treepolicy$ is parameterized by a neural network). To address this issue, we nest a sampling procedure within a given Monte Carlo rollout sample, effectively Monte Carlo sampling a Monte Carlo rollout of our tree policy. More precisely, we stochastically approximate $\tree_{\treepolicy}(\satinstance)$ by using Knuth samples \cite{Knuth1975}.
We first state the general version of Knuth's theorem for estimating the total weight of a weighted $k$-ary tree with a random path through the tree.
\begin{thm}[\citet{Knuth1975}]
\label{thm:knuthsamp}
Let $\nodes$ be the set of nodes in a weighted $k$-ary tree $\tree$ where each node $\node\in \nodes$  has weight $\weight(\node)$. The total weight of this tree is $\weight_{\tree} = \sum_{\node\in\nodes}\weight(\node)$.  Let $(n_{0},\ldots, n_{\length})$ be a path of nodes through the tree from root to leaf that is chosen uniformly at random.
Then, 
$\weight_{\tree} = 
\Ex[(n_{0},\ldots n_{\length})]{\sum_{i=0}^{\length} k^{i}\weight(n_{i})}$.
\end{thm}
In the case of binary trees where all nodes have weight $1$, a simple corollary holds.

\begin{cor}
\label{cor:knuth}
Let $\length_{\kpath}$ be the length of a path $\kpath$ $(n_{0},\ldots, n_{\length_{\kpath}})$ where $\forall_i, w(n_{i})=1$ and sampled uniformly at random from a binary tree $\tree$ with size $\size_\tree$. Then, 
$\size_{\tree} = 
\Ex[P]{[2^{\length_{\kpath+1}}-1]}$.
\end{cor}
\begin{proof}
When $w(n_{i})=1$, $s_{T} = E_P{[\frac{k^{\ell_{P} }-1}{k-1}]}$. This is true because $\sum_{0}^{n-1} k^{i} =\frac{k^{n}-1}{k-1}$. Plugging in $k$=2 (binary tree in UNSAT), we get $s_{T} = E_P{[2^{\ell_{P} }-1]}$.
\end{proof}

For the tree $T$ produced by $\treepolicy$, we can use Theorem \ref{thm:knuthsamp} to get an unbiased estimate of
$\tree_{\treepolicy}(\satinstance)$. In the context of a DPLL solver, a Knuth sample amounts to replacing a complete traversal of the binary tree of all \emph{True} / \emph{False} assignments with a path through the tree where assignments are chosen uniformly at random. We can take the length of the resulting path, $\length$, and update the tree size estimate of each node at depth $\depth$ with $2^{\length-\depth}-1$ for the corresponding decision by $\alpha$. The average of these estimates across a set of paths yields an unbiased approximation of the tree size. 

\subsection{Bounding Policy Evaluations}
\label{sec:bounding}
Knuth Synthesis is an offline procedure for training a variable selection policy $\policy$. At test time, we want proofs of unsatisfiability rather than estimates of tree size, and so cannot use Knuth samples. 
We represent $\policy$ using a deep neural network, which is far more expensive to evaluate than conventional tree-search heuristics. This raises the risk that the computational cost of evaluating $\policy$ will exceed its benefits via proof tree size reductions, and indeed makes this risk a certainty for small enough subproblems, such as those for which existing solvers are faster than single network calls. Any online policy must therefore have a procedure for constraining the nodes at which $\policy$ will be queried. 
As we saw earlier, decisions become more important the closer they are to the root of an UNSAT search tree. %Since our policy is expensive and we cannot afford to use it for every decision, 
We thus apply $\policy$ at states having depth $\leq\length$, after which we either (1) at both test and training time, call a ``subsolver'' (a pre-existing variable selection policy); or (2) at training time, sometimes instead call a value network that more cheaply approximates the tree size of this fixed policy (see ``Rollout Policy'' for details). %\footnote{Another reason to use node expansion is that it can lower the variance of the rewards below a given state. However, we noticed empirically that the variance does not grow significantly with sample length within the range of number of decisions before we call a subsolver. Therefore, the variance reduction does not outweigh the benefit of removing expensive subsolver calls.}

Although we could potentially afford to explore larger $\length$ values at training time, we do not do so; this ensures that the training phase identifies a policy that will leverage the subsolver appropriately at test time. To determine the appropriate tree size estimation, we add a weight of $\tree_{\policy_{sub}}(\satinstance^*)$ to any node representing state $\satinstance^*$ at which we call the subsolver policy $\policy_{sub}$. This represents the additional tree size incurred from the subsolver call at this node. For a given Knuth sample of length $\length$ that terminates at $\satinstance^*$, we can use the weighted version of Knuth's theorem to update the value of a node at depth $\depth$ along the path with 
\begin{kequation}
\label{eqn:knuth_update}
2^{\length-\depth}\tree_{\policy_{sub}}(\satinstance^*) + 2^{\length-\depth}-1.
\end{kequation}

\subsection{Implementation} 
 Beyond changes to avoid exponential evaluation costs, Knuth Synthesis has a few other noteworthy components. We %leverage ideas from AlphaZero where appropriate and 
alter the Action Selection step to account for the tree size cost function that scales with node depth; we adapt the Simulation step to account for an unreliable value network and a fixed policy; we use a model architecture to respect the structural invariances of SAT; we use a graph-based state-transition data structure to improve sample efficiency; and we share policy priors with child nodes to speed up lookaheads. Find pseudocode in Appendix \ref{app:pseudocode}.
\subsubsection{Tree Policy $\treepolicy$}
At every state $\state$, $\treepolicy$ returns action 
\begin{kequation}
\action = \argmin_{\action'} (Q(\state,\action') - Q_{\depth}U(\state,\action') ).
\end{kequation}
$Q(\state,\action)$ is the cost estimate of action $\action$ at state $\state$, $Q_{\depth}$ is the running average tree size of nodes at depth $\depth$, which is used to calibrate confidence intervals across different depths, and 
\begin{kequation} 
U(\state,\action) = \cpuct P(\state,\action)\frac{\sqrt{\sum_{\action'} N(\state,\action')}}{1 + N(\state,\action)}
\end{kequation}
is the corresponding confidence interval \citep{Rosin2011,Silver2016}. The confidence interval is parameterized by $N(\state,\action)$, the number of lookaheads that branch on action $\action$ at state $\state$,  $\cpuct$, a constant that controls exploration, and $P(\state,\action)$, a prior distribution over the actions for a given state, predicted by the policy network.  $N(\state,\action)$ is initialized to 1 and $Q(\state,\action)$ is initialized with the tree size of the first lookahead at that state. This provides an unbiased measure of the performance of our incumbent policy (i.e., the first sample is exactly the neural network policy), which we seek to improve upon. This is similar to AlphaZero's choice of initializing $Q$ to 0, which is the baseline they aim to improve upon.  Because $Q(\state,\variable)$ scales exponentially with the depth of $\state$, we introduce $Q_{\depth}$ to calibrate confidence intervals. Since $U(\state,\variable)$ is independent of the scale of costs (it is only a function of counts), we calibrate each depth with $Q_{\depth}$ to keep $U(\state,\variable)$ at the same scale as $Q(\state,\variable)$. Otherwise,  the choice of the $\cpuct$ would trade off tight confidence intervals at shallow nodes against loose confidence intervals at deeper nodes. We compute $Q_{\depth}$ online as the running average of tree sizes encountered at depth $\depth$. 

\paragraph{Equivariant Architecture}
SAT problems vary in size and are invariant to permutations of their clauses and variables within clauses. We enforce these invariances using a permutation equivariant architecture in which permutations of the input guarantee a corresponding permutation in the output \citep{Hartford2018}. A beneficial side effect of this architecture is that it can handle input matrices of any size. 
\paragraph{Sharing Prior with Child Nodes}
Calls to the policy network tend to dominate running time, so we save time by only computing $P(\state,\action)$ once at the root of the MCFS forest and passing down the prediction to its child nodes. A child state has a subset of the action space of its parent state. %When we reach a new 
For child state $\state'$, we set $P(\state',\action) \leftarrow P(\state,\action)$, and renormalize for the subset of actions remaining in $\state'$.

\paragraph{Directed Acyclic Graph as Forest Search Data Structure} 
We improve sample efficiency by changing our forest search data structure from a tree to a graph, which leverages the fact that states are reached independently of the order of previous decisions \cite{Czech2020}. This change ensures information sharing across different paths that lead to the same state.

\subsubsection{Step Policy $\steppolicy$}
An MCFS Step policy $\steppolicy$ controls how much of the proof tree to evaluate in a rollout. In order for $\steppolicy$ to implement Knuth sampling, it should play actions along a single path through the rollout tree, choosing actions uniformly at random. We must therefore ensure that for every pair of newly visited states ($s^{L},s^{R}$), $\steppolicy$ only plays an action at one of them. If $\steppolicy$ is called at a state $s$ and its sibling $s^{sib}$ has yet to be processed, then $\steppolicy$ chooses between $\treepolicy(s,)$ and $\emptyset$ with uniform probability. If $s^{sib}$ has already been processed, then $\steppolicy$ returns the opposite of the decision made at $s^{sib}$. $\steppolicy$ will also return $\emptyset$ if it encounters a state $s$ at depth $\length$ or it encounters a leaf node ($s$ s.t. $l(s)=0$) i.e., a conflict in the underlying SAT problem.

\subsubsection{Rollout Policy $\rolloutpolicy$}
At each state $s$, $\rolloutpolicy$ can be described via two cases: (1) if $s$'s sibling $s^{sib}$ played an action (i.e. $\steppolicy(\state^{sib})\neq \emptyset)$ then $\rolloutpolicy(\state)$ returns the cumulative reward at $s^{sib}$, ensuring that the reward estimates at any given node reflect the correct Knuth estimate; (2) no action was played at $s$ or its sibling because  $s$ is at depth $\length$. In this case our rollout policy $\rolloutpolicy$ is to call the subsolver at a state $\satinstance^{*}$ at depth $\length$ of our Knuth sample. Rather than explicitly solving the subproblem at depth $\length$, we use a value network to access a much cheaper reward signal. This network is trained from an initial batch of subsolver calls and then retrained as we collect new batches of data, so its accuracy varies across time steps of a training run. When the output of the network is too unreliable, its signal about the true number of decisions may be insufficient for policy learning. We address the issue of an unreliable network by randomly deciding between calling the subsolver or the value network, with the probability depending on an online estimate of the value network’s accuracy. Specifically, we track the mean multiplicative error $\epsilon$ of our value network over time by querying the value network with every subsolver call. For a user-defined accuracy threshold parameter $t$ (we use $t = 0.5$), we sample the value network with probability $1- \min(1, \epsilon / t)$, so that the probability of calling the subsolver halves as the error halves.

\section{Experimental Setup} 
\label{app:experiments}

\subsection{Benchmarks} 
We targeted instance distributions that are well known and difficult for modern SAT solvers. We controlled instance size so that (1) the size of the action space was similar to Go and (2) solving required $\approx$ 100,000 decisions. We did not consider industrial SAT Competition instances, as they often contain millions of variables with state spaces significantly larger than any deployed MCTS application  of which we are aware. We evaluated our approach on three distinct distributions: (1) a canonical random distribution: uniform random 3-SAT at the solubility phase transition (\texttt{R3SAT}), (2) a notoriously difficult crafted distribution (\texttt{sgen} \cite{Spence2010}), and one real-world application: station repacking problems from the 2016 FCC incentive auction (\texttt{satfc} \citep{Frechette2016}). For \texttt{R3SAT}, we trained on 300-variable instances, for which calling a subsolver was quick ($\approx$ 1 second solving time), and filtered out satisfiable instances for training and testing.  
To evaluate upward size generalization, we set aside 100 test instances at both our training size of 300 variables as well as at 350, 400, and 450 variables.   
For \texttt{sgen}, we trained on 65 variables; to evaluate upward size generalization we set aside 100 test instances at both our training size of 65 variables as well as at 75, 85, and 95 variables.  For \texttt{satfc}, we trained on small ($<$2000-variable) instances from localized regions of the U.S. interference graph, and filtered out satisfiable instances. We set aside 100 test instances of our training distributions. For full details, see Appendix \ref{app:benchmarks}.

\subsection{Baselines}
Since Knuth Synthesis learns DPLL policies, we sought to evaluate against other purely DPLL solvers rather than to make apples-to-oranges comparisons to clause learning (CDCL) solvers. We ultimately aim to integrate our MCFS ideas into the much richer design space of CDCL solvers, but anticipate that this will require going beyond the tree MDP formalism.
%Since we did not consider the much bigger design space available in modern clause learning (CDCL) solvers, our experiments aimed to evaluate whether we could synthesize strong solvers within the DPLL framework rather than to show state-of-the-art performance. We see this as a key step before we integrate powerful algorithmic concepts into MCFS such as clause learning and restarts. 
To make our comparisons as fair as possible, we set our baseline and subsolver to be the same: \texttt{kcnfs} \citep{Dequen2003}, which is a pure-DPLL solver and is specifically designed for and is among the strongest solvers for \texttt{R3SAT}. We used its most recent competition submission at the 2007 SAT Competition \citep{SATcomp2007}, where it won the silver medal in the Random UNSAT track and had previously won the gold medal in the 2005 Random UNSAT track. The lookahead \texttt{march} solver \cite{Heule2009} is $\approx$10\% faster but we chose \texttt{kcnfs} because it is a pure-DPLL solver. \texttt{kcnfs} is a much stronger baseline than the minisat baseline which \citet{Kurin2020} compare to (two orders of magnitude faster on their benchmarks). We also evaluated a \texttt{uni+kcnfs} baseline on each dataset, where we replaced Knuth Synthesis neural network calls with uniform-at-random decisions and called the \texttt{kcnfs} solver for subproblems at the same user-defined depth. We tried a purely random policy without calling \texttt{kcnfs} as a subsolver on 65-variable \texttt{sgen}, which led to poor performance $40\times$ slower than \texttt{kcnfs}. %Thus, we did not evaluate purely random policies beyond this dataset. We also evaluated a \texttt{JW+kcnfs} baseline where we replaced neural network calls with the well known Jeroslav-Wang DPLL heuristic \citep{jeroslow1990solving}, which was  $2.6\times$ and $3.7\times$ slower than \texttt{kcnfs} on \texttt{sgen-65} and \texttt{R3SAT-300}, respectively.

\subsection{Knuth Synthesis}

We integrated our Knuth Synthesis algorithm into the CDCL solver \texttt{Maple\_LCM\_Dist\_\\ChronoBT} \citep{Ryvchin2018}, which won the 2018 SAT Competition and is based on the MiniSAT framework 
\citep{Een2003}. We removed all clause-learning components so that the solver ran pure DPLL search. We trained our neural networks with PyTorch \citep{Paszke2019} in Python and ported them to our C++ solver using tracing. Find our code here: \url{https://github.com/ChrisCameron1/MCFS}.

Two important hyperparameters were (1) the constant for the level of exploration $\cpuct$ and (2) the number of lookaheads $\lookaheads$. Using a coarse grid search, we selected $\cpuct=0.5$ and $\lookaheads=100,000$, which found the best policies within a 48-hour window. We chose $\length =$ 5, 6, and 8 for the \texttt{satfc}, \texttt{R3SAT} and \texttt{sgen} distributions respectively, based on the average tree size of Knuth Synthesis after 48 hours for values between 2 and 10 over a few instances. This $\length$ parameter trades off between (1) flexibility of the policy (neural net makes more decisions) and (2) size of the search space. There is a sweet spot where the search space is large enough to find an improving policy but small enough that we can learn that policy in a reasonable amount of time. Fewer variables in a problem means we can search deeper for an equivalently-sized search space so that likely explains why we observed better results for larger $\length$ as the number of variables increased: 5 for \texttt{satfc} ($\approx$1000 vars), 6 for \texttt{R3SAT} (300 vars), 8 for \texttt{sgen} (65 vars). We search deepest in \texttt{sgen} where there are fewest variables. The best $\length$ will also depend on the amount of computation we allocate for Knuth Synthesis; with more compute and/or a more efficient MCFS algorithm, we could search considerably deeper.
We made decisions with Knuth Synthesis until depth $\length$; each MCFS decision yielded a training point consisting of a (state, policy vector, $Q$-value) triple, where the policy vector represents normalized action counts from MCFS. 

Instead of uniformly random sampling true/false variable assignments during Knuth samples, we tried to reduce variance by importance sampling based on value network estimates. We did not observe any variance reduction likely because our value network was not sufficiently strong.

Before a good policy network is learned, Knuth Synthesis tends to be less efficient. We pretrained our policy and value networks by running Knuth Synthesis with 10,000 lookaheads on 1,000 instances for \texttt{R3SAT} and \texttt{sgen}. We ran one iteration of Knuth Synthesis with 10,000 lookaheads on 1,000 instances to further improve the policy and value network, and then a final iteration with 100,000 lookaheads on 2,000 instances to train our final model. Pretraining runs of Knuth Synthesis took approximately 24 hours per instance and were respectively run on 300-variable and 55-variable problems from \texttt{R3SAT} and \texttt{sgen}. The subsequent iterations took approximately 48 hours per instance. On \texttt{R3SAT} and \texttt{sgen}, they were respectively run on 300-variable and 65-variable problems. For \texttt{satfc}, we pretrained with 10,000 lookaheads on 441 instances and ran a final iteration using the prior %on the same 441 instances 
with 100,000 lookaheads. The parameters and architecture described in this paper are only for the last iteration of Knuth Synthesis. We made several minor improvements across iterations.

\subsection{Model Training} 
We used the exchangeable architecture of \citet{Hartford2018}. We represented a CNF SAT instance with $n$ clauses and $m$ variables as an $n\times m\times 128$ clause-variable permutation-equivariant tensor, where entry $(i, j)$ is $t_{v}$ if the true literal for variable $i$ appears in clause $j$, $f_{v}$ if the false literal for variable $i$ appears in clause $j$, and 0 otherwise. $t_{v}$ and $f_{v}$ are 128-dimensional trainable embeddings representing the true and false literal. Following \citet{Hamilton2017}, we also added a node degree feature to every literal embedding. We instantiated the permutation-equivariant portion of the exchangeable architecture as four exchangeable matrix layers with $512$ output channels, with leaky RELU as the activation function. We mean pooled the output to a vector, with each index representing a different variable. We experimented with attention pooling in the last exchangeable layer but did not observe any improvements. 

We added three feed-forward heads: a policy head, a $Q$-value head, and a value head. The policy and $Q$-value heads both had two feed-forward layers with $512$ channels and a final layer that mapped to the single output channel. The value head was the same, except there was a final mean pool that output a single scalar. The policy head and the $Q$-value head were trained to predict the normalized counts and $Q$-values of Knuth Synthesis, respectively. We optimized cross-entropy loss for both the policy and $Q$-value head. The purpose of the $Q$-value head was as an auxiliary task to help learn a better shared representation for the policy head; we observed a 5\% reduction in tree size after adding the $Q$-value head. We trained the value network with mean-squared error against the $\log_{2}$ tree size of subsolver calls at leaf nodes. The value head was not backpropagated through the exchangeable layers. Using a shared representation was important for training the value head; loss tended to double when using a network trained with only a value head. For a given depth $d$, MCFS makes $2^d$ decisions; therefore, we received exponentially more data points at deeper depths where decisions are less important. We experimented with exponentially upsampling nodes inversely proportional to their depth but observed no performance gains. We used the Adam optimizer \cite{Kingma2014}, with a learning rate of $0.0001$ and a batch size of 1. We evaluated mean tree size on a held-out validation to select a model. Online, we branched on the argmax variable from our neural network prediction. 

\subsection{Computing Resources}
We ran our model training and solver benchmarking experiments on a shared cluster with A100 GPUs, equipped with 32 2.10GHz Intel Xeon E5-2683 v4 CPUs with 40960 KB cache and 96 GB RAM each, running openSUSE Leap 42.1 (x86\_64). For Knuth Synthesis runs, we used a large shared cluster of CPU nodes with 1109 nodes, equipped with 64 2.40 GHz 2 x AMD Rome 7532 with 256 MB of L3 cache. Each Knuth Synthesis run was allocated 16 GB of memory and a maximum of 48 hours for \texttt{R3SAT} and \texttt{sgen} and 72 hours for \texttt{satfc}. Each benchmarking run was allocated 8 GB of memory.

\section{Results} 
\label{sec:exp-results}

\begin{table}[t] 
\small
\centering
\setlength{\tabcolsep}{1.0pt}.
\begin{tabular}{lll@{\hskip 0.2cm}rrr@{\hskip 0.2cm}rrr}
\toprule
&  & & \multicolumn{3}{c}{\bf Tree size (1000s) } &  \multicolumn{3}{c}{\bf CPU+GPU time (s) } \\
 & Size & In/Out & {\fontsize{8pt}{10pt}\selectfont\ttfamily uni+kcnfs} & \texttt{kcnfs} & Ours & \texttt{uni+kcnfs} & \texttt{kcnfs} & Ours \\

\midrule
 \texttt{R3SAT} & 300 & In & 44.3 & 8.9 & {\bf 8.6 (2\%)} & 15.7  & {\bf 1.1} & 10.8 \\
  & 350 & Out & 215.7 & 43.3 & {\bf 42.7 (1\%)} & 59.5 & {\bf 6.1} & 15.8\\
 & 400 & Out &1,103.5 & 226.9 & {\bf 223.5 (2\%)} & 289.1 & {\bf 30.8} & 36.5 \\
  & 450 & Out &5,207.9 & {\bf 989.7} & 994.9 & 1591.1 & {\bf 168.9} & 173.4 \\
\midrule
 \texttt{sgen} & 65 & In & 158.2 & 162.3 & {\bf 132.2 (23\%)} & 8.5 & {\bf 2.1} & 8.0\\
  & 75 & Out & 1,799.3 & 1,792.3 & {\bf 1,594.0 (12\%)} & 26.6 & {\bf 23.1} & 26.1 \\
 & 85 & Out & 8,932.9 & 8,874.8 & {\bf 8,156.2 (9\%)} &  115.0 & 114.4 & {\bf 105.6(8\%)}\\
  & 95 & Out & 98,534.0 & 97,979.7 & {\bf 92,407.9 (6\%)} & 1214.7 & 1272.2 & {\bf 1,178.7 (8\%)}\\
 \midrule

 \texttt{satfc} & &  In & 631.3 &  250.1 & {\bf 67.1 (372\%)} &  24.8 & 8.3 & {\bf 6.5 (128\%)}\\
  
\bottomrule
\end{tabular}
\caption{Mean tree size (1000s of nodes) and running time (CPU+GPU seconds) comparing Knuth Synthesis (Ours), \texttt{kcnfs}, and \texttt{uni+kcnfs} over 100 test instances from each distribution. Reductions in decisions and running time are relative to \texttt{kcnfs}. In/Out denotes whether the benchmark was in or out of the training distribution.%\cc{Report computation in dollars? Correct between GPU and CPU cost.}
}
\label{tab:results}

\end{table}

We evaluated (1) search tree size and (2) running time for Knuth Synthesis  against the two baselines on each of our  instance sets; the full results are presented in Table \ref{tab:results}.  We measured running time as the cumulative CPU and GPU time with \texttt{runsolver} \citep{Roussel2011}. %; we set aside 100 test instances for each benchmark. 
Overall, we reduced tree size on each of the three training distributions as well as 5/6 upward-size generalization distributions.  This led to running time improvements on 3/9 datasets where the neural network overhead was manageable. 
\texttt{R3SAT} was the most challenging benchmark as \texttt{kcnfs} is an extremely strong solver specialized to this distribution; it far surpassed random branching on top-level decisions (4--5$\times$ reduction in tree size and walltime)  and it was unclear whether it could be improved upon. Despite the strength of \texttt{kcnfs}, we squeezed out performance improvements of 1-2\% in average tree size over \texttt{kcnfs} on up to 400 variables. Given the overhead of our neural network calls, these reductions in tree size did not lead to improvements in running time. For \texttt{sgen}, \texttt{kcnfs} was also very strong (see Appendix \ref{app:expanded_results}) however there is greater scope for improvement since it was not optimized for \texttt{sgen}. %We trained on 65 variables and evaluated at 65, 75, 85, and 95 variables.
%On \texttt{sgen}, there is no known specialized solver and therefore at least somewhat greater scope for improvement, however \texttt{kcnfs} was still better than the best SAT competition solver. 
%We observed that \texttt{kcnfs} usually found only marginally smaller trees than \texttt{uni+kcnfs}, but it was the best existing solver we were able to identify. %. However, we were unaware of any solver that performs well on this distribution. 
We reduced average tree size over \texttt{kcnfs} on our training distribution (65 variables) by 1.23$\times$. Our model trained on the 65-variable distribution generalized well to larger problem sizes; it reduced tree size even on 95 variables, which took $\approx 20$ minutes to solve ($700\times$ more difficult). Our solver incurred a roughly constant overhead that prevented us from improving running time on 65 variables and 75 variables. %However, asymptotic behaviour is considered a better indicator of performance by the SAT community. 
We were able to improve the running time over \texttt{kcnfs} by 8\% on 85 variables and 8\% ($\approx1.5$ minutes faster) on 95 variables. Even without using a GPU for model queries, we improved running time by 8\% at 95 variables (constant policy-query overhead is swamped by tree-search time at higher running times). For \texttt{satfc}, evaluating against \texttt{kcnfs} allowed us to evaluate the (more realistic) scenario where there was no strong hand-tailored baseline. Such settings offer the likelihood of large scope for branching policy improvements; we saw the question of whether Knuth Synthesis discovered such policies as an important test. \texttt{kcnfs} ran $\approx 3\times$ faster than \texttt{uni+kcnfs}. We reduced tree size by $3.72\times$ and reduced running time by $1.28\times$ over \texttt{kcnfs}. Find expanded results in Appendix \ref{app:expanded_results} including an experiment that shows the infeasibility of an existing RL approach on our datasets.

 \subsection{Knuth Speed Up}
\begin{figure}[b]
    \centering
    \includegraphics[width=\columnwidth,trim={0 5pt 0 10pt}]{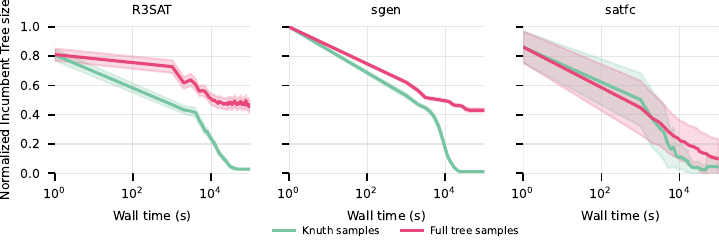}
    \caption{Convergence rate improvements with Knuth samples. We normalize tree size across instances by the per-instance minimum and maximum tree size ever observed. Shaded regions represent 95\% confidence intervals.}
    \label{fig:knuth_convergence}
\end{figure}

We evaluated how much quicker we could find good policies using Knuth samples. Each $\depth$-bounded Knuth sample is in expectation $2^\depth$ cheaper than evaluating the full proof tree but the variance of the estimates can be high for unbalanced proof trees. We hoped variance  would be sufficiently small that we would need much fewer than $2^\depth$ samples to distinguish between good and bad policies, allowing us to more quickly move to better parts of the policy space. Figure \ref{fig:knuth_convergence} shows average tree size over time comparing Knuth Synthesis to full-tree evaluation for our three training benchmarks. We normalized tree size across instances by the per-instance minimum and maximum tree size ever observed. In all cases, there was a clear efficiency improvement using Knuth samples. On \texttt{R3SAT} and \texttt{sgen}, Knuth sampling reached a given tree size approximately an order of magnitude more quickly. The improvement on \texttt{satfc} was substantial but not as pronounced; we hypothesized this was because the \texttt{satfc} proof trees tended to be less balanced, which resulted in higher-variance Knuth estimates.

\section{Conclusions and Future Work} 
\label{sec:discussion}
We presented MCFS,  a class of RL algorithms for learning policies in tree MDPs. We introduced an MCFS implementation called Knuth Synthesis that approximates tree size with Knuth samples to avoid the prohibitive costs of exact policy evaluation that existing RL approaches suffer from. We matched or improved performance over a strong baseline in a diverse trio of distributions, tackling problems within these distributions that were two orders of magnitude more challenging than those in previous studies. In future work, we first would like to generalize MCFS to CDCL solvers. Most high-performance industrial solvers use the CDCL algorithm, which adds a clause-learning component to DPLL to allow information sharing across the search tree. This information sharing means there is no straightforward way to encode the problem as a tree MDP. Approximating the size of a CDCL search tree with Knuth samples is non-trivial: unlike DPLL, single paths from the root to leaves cannot be evaluated independently to approximate tree size. The only obvious way to apply MCTS for CDCL policies is for a path to represent a full exploration through the proof tree, which would make the search space exponentially larger. Next, we would like to scale MCFS to industrial-size problems with millions of variables. There are two main bottlenecks to achieving this: (1) GPU memory constraints and (2) efficiently searching a much larger action space. Memory requirements scale linearly with problem size and our resources restrict us from training beyond 10,000 variables; we would like to explore model parallelism, memory-efficient gradient approximations, and smaller state representations. We also would like to reduce the action space through learning an action representation and collapsing similar actions together. Finally, we would like to explore MCFS for satisfiable instances, which breaks our tree MDP property. Finding a small proof tree may be complementary to finding a SAT solution; if we can quickly prove that most of the search space does not have a solution, we can search for a solution in a more restricted space. We are interested in exploring a hybrid idea that simultaneously looks for SAT solutions at the same time being rewarded for shrinking the search space. 

% \paragraph{Training on Other Targets to Learn Better Representations}

% The biggest insight we gained from training our models was the importance of leveraging training on one concept to build a representation that could better learn another concept. We could not learn an efficient value network without jointly training the policy network, and we were also able to learn a better policy network from jointly predicting $Q$-values. We would like to explore other concepts that we could jointly train with to improve our policy and value networks further. One idea is to additionally supervise with the UNSAT core, which was shown by \citep{Selsam2019} to be an effective heuristic.

\section*{Acknowledgments}
We thank the reviewers for all the constructive feedback. This work was funded by an NSERC Discovery Grant, a DND/NSERC Discovery Grant Supplement, a CIFAR Canada AI Research Chair (Alberta Machine Intelligence Institute), a Compute Canada RAC Allocation, awards from Facebook Research and Amazon Research, and DARPA award FA8750-19-2-0222, CFDA \#12.910 (Air Force Research Laboratory). We thank Greg d'Eon for many useful conversations, especially one infamous AlphaGo tutorial.

\bibliographystyle{splncs04nat}
\bibliography{references}

\appendix

\newpage
The appendix is divided into three sections: Pseudocode, Expanded Results, and Benchmark Details.

\section{Pseudocode}
\label{app:pseudocode}

\subsection{Knuth Synthesis}

We now provide pseudocode for Knuth Synthesis' implementation of the three pieces of MCFS: tree policy $\treepolicy$ (Algorithm \ref{alg:treepolicy}), step policy $\steppolicy$ (Algorithm \ref{alg:steppolicy}), and rollout policy $\rolloutpolicy$ (Algorithm \ref{alg:rolloutpolicy}). Parameters/inputs specific to the Knuth Synthesis implementation are declared in the first line of each algorithm.

\begin{minipage}{0.7\textwidth}
\begin{algorithm}[H] 

\caption{$\alpha(s,c,v)$}
 %\hspace*{\algorithmicindent} 

\begin{algorithmic} 
\STATE \textbf{Input:} $c_{PUCT}$, depth calibration $Q_{d}$, prior $p$
\STATE  $\forall a, u_{\action}= c_{PUCT}p_\action\frac{\sqrt{\sum_{\action'} c_{\action'}}}{1 + c_\action}$
\STATE return $\argmin_{\action'} (v_{\action'} - Q_{d}u_{\action'})$
\end{algorithmic}
\label{alg:treepolicy}
\end{algorithm}
\end{minipage}

\begin{minipage}{0.7\textwidth}
\begin{algorithm}[H] 
\caption{$\steppolicy(\rollout,\action,\state$)}
 %\hspace*{\algorithmicindent} 

\begin{algorithmic} 
\STATE \textbf{Input:} rollout depth $\length$
\IF{Depth($\state$,$\rollout$) > $\length$ or $l(\state)=0$}
\STATE return $\noaction$
\ENDIF
\STATE $\sibling \leftarrow$ sibling($\state$)
\STATE \emph{/* Case 1: If sibling processed, take opposite */}
\IF{processed($\rollout, \sibling$)}
    \STATE $\action' \leftarrow \steppolicy_{cache}(\sibling)$
    \IF{$\action' = \noaction$ }
        \STATE step $\leftarrow$ $\action$
    \ELSE
        \STATE step $\leftarrow$ $\noaction$
    \ENDIF
\ELSE    
    \STATE \emph{/* Case 2: Flip a coin by Knuth */}
    \IF{Random draw with p=0.5 is True}
        \STATE step $\leftarrow$ $\action$
    \ELSE
        \STATE step $\leftarrow$ $\noaction$
    \ENDIF
\ENDIF
\STATE return step
\end{algorithmic}
\label{alg:steppolicy}
\end{algorithm}
\end{minipage}

\begin{minipage}{0.7\textwidth}
\begin{algorithm}[H] 
\caption{$\rolloutpolicy(\state)$}
 %\hspace*{\algorithmicindent} 

\begin{algorithmic} 
\STATE \textbf{Input:} running multiplicative error $\epsilon$, accuracy threshold $t$, subsolver calls $n$, value net $V$, subsolver policy $\policy_{sub}$
\STATE $p \leftarrow 1- (\epsilon / t$)
\IF{Random draw with prob $p$}
    \STATE return $V(\state$)
\ELSE
    \STATE $m \leftarrow | log(T_{\subsolver}(\state)) - log(V(\state)|$ \emph{/* multiplicative error */}
    \STATE $\epsilon \leftarrow  \epsilon + \frac{(m - \epsilon)}{n+1} $ \emph{/* $O(1)$ update */}
    \STATE $n \leftarrow n+1$
    \STATE return $T_{\subsolver}(\state)$
\ENDIF
\end{algorithmic}
\label{alg:rolloutpolicy}
\end{algorithm}
\end{minipage}

\subsection{DPLL}
\label{app:dpll}
The DPLL algorithm (Algorithm \ref{alg:dpll}) takes in some branching policy $\policy$, which outputs a variable given the SAT instance. DPLL then recursively calls itself with assigning that variable to true and false. Given a variable assignment , we call the transition function $\tau_{DPLL}$ to find the new simplified $\satinstance$. $\tau_{DPLL}$ makes calls to two subroutines.

\paragraph{Unit Propagation} Unit propagation finds all unit clauses i.e., clauses which contain only a single unassigned literal. It then removes all other clauses which contain that literal; from all clauses that contain the literal's complement, it removes the literal's complement.

\paragraph{Pure Literal Assignment} Pure literal assignment finds all literals such that their complements are not present in the SAT instance, which are known as pure literals. It then removes every clause that contains a pure literal because those can be trivially satisfied.

\subsubsection{Tree MDP definition for DPLL UNSAT}
See Definition \ref{def:treemdp} for the definition of a tree MDP. An action $\action$ is a choice of variable to branch on. The reward function is $r(s_{i}) = -1 \forall i$, as we are trying to minimize the number of nodes. A leaf node i.e., $l(s_i) = 1$ is any state $s_i$ where DPLL returns \emph{False} or \emph{True}. $s_{ch_{i}}^{L}$ will be $s \wedge (\action_i = 0)$ and $s_{ch_{i}}^{R}$ will be $s \wedge (\action_i = 1)$. Note that for satisfiable instances, this last property does not necessarily hold. We may only need to transition to $s_{ch_{i}}^{L}$ or $s_{ch_{i}}^{R}$ to solve the problem, if either of the subproblems is SAT. It is not correct to model SAT (as opposed to UNSAT) with tree MDPs.

\begin{algorithm}
\caption{DPLL}
 \hspace*{\algorithmicindent} \textbf{Input:} SAT instance $\satinstance$, policy $\policy$, assignment $x_{i}$
 
\begin{algorithmic} 
% \Procedure{OuterLoop}
% \STATE $V \leftarrow \{\}$
% \WHILE{iterations less than $H$}
% $\mathrm{Rollout}(\satinstance)$
\STATE $\tau_{DPLL}(\satinstance,x_i)$
\IF{$\satinstance$ is empty}
\RETURN \emph{True}
\ENDIF
\IF{$\satinstance$ contains an empty clause}
\RETURN \emph{False}
\ENDIF
\STATE $\variable \leftarrow \policy(\satinstance)$
\RETURN $\mathrm{DPLL}(\satinstance, \policy, \variable=0)$ OR $\mathrm{DPLL}(\satinstance, \policy, \variable=1)$
\end{algorithmic}
\label{alg:dpll}
\end{algorithm}

\begin{algorithm}
\caption{$\tau_{DPLL}$}
 \hspace*{\algorithmicindent} \textbf{Input:} SAT instance $\satinstance$,\\ \hspace*{4\algorithmicindent} assignment $x_{i}$
\begin{algorithmic} 
% \Procedure{OuterLoop}
% \STATE $V \leftarrow \{\}$
% \WHILE{iterations less than $H$}
% $\mathrm{Rollout}(\satinstance)$
\STATE $\satinstance \leftarrow \satinstance \wedge x_{i}$ 
\STATE $\satinstance \leftarrow \mathrm{UnitPropagation}(\satinstance)$
\STATE $\satinstance \leftarrow \mathrm{PureLiteralAssign}(\satinstance)$
\RETURN $\satinstance$
\end{algorithmic}
\end{algorithm}

\section{Expanded Results}
\label{app:expanded_results}

\subsection{Comparisons to past work}

Past work has never scaled beyond 1000 decisions during training episodes and have no mechanism for approximating policy evaluations, therefore we hypothesized that these methods would be infeasible at scale. To test this, we evaluated our approach against Graph-Q-SAT \cite{Kurin2020}, the RL-for-SAT approach that is closest to our own. We could not find code to compare to \citet{Lederman2019} and we could not get \citet{Vaezipoor2021}'s code working, however we think Graph-Q-SAT performance is a good indicator of the success of other existing approaches.  \citet{Vaezipoor2021}'s approach is a black-box rather than RL method but also relies on exact policy evaluation. \citet{Kurin2020} trained Graph-Q-SAT on 50-variable \texttt{R3SAT} problems.  We evaluated whether we could train Graph-Q-SAT on larger-sized problems from the same distribution up to 250 variables, restricting to unsatisfiable instances. For each size at increments of 50 variables, we trained a Graph-Q-SAT model for five days with a decisions cap of 100,000 over a training set of 1000 instances. We evaluated validation performance after every 1000 training episodes and evaluated the model with best validation performance. Graph-Q-SAT is integrated with \texttt{minisat} \cite{Een2003}; for a fair comparison, we trained our method using the same \texttt{minisat} as our rollout solver and all other parameters left unchanged. We could not easily benchmark Graph-Q-SAT with \texttt{kcnfs} (which is orders of magnitude faster), since they are tightly integrated with into minisat with a customized SWIG interface bridging C++ and python. We reported number of decisions (no runtimes) since Graph-Q-SAT calls out to \texttt{minisat} in a very inefficient way that makes their method appear artificially slower.

\begin{figure}
\includegraphics[width=\textwidth]{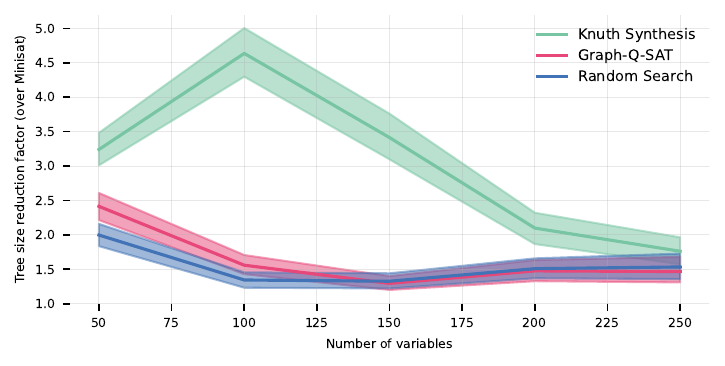}
\caption{Graph-Q-SAT vs. Knuth synthesis on unsatisfiable R3SAT problems with increasing size from 50 to 250 variabels at 50 variable increments. Performance is measured as the multiplicative factor of improvement over \texttt{minisat}.}
\label{fig:qgraph-comparison}
\end{figure}

Figure \ref{fig:qgraph-comparison} shows performance on a held-out test set for each problem size. Performance is measured as the multiplicative reduction factor in tree size over the \texttt{minisat} baseline. Knuth Synthesis is better at every problem size including making over 2$\times$ fewer decisions at 100 variables. Graph-Q-SAT still performs quite well. It improved over \texttt{minisat} by 2.5$\times$ at 50 variables and by 1.5$\times$ on 100-250 variables. We were confused how Graph-Q-SAT could make learning progress when these numbers of decisions were required (100,000 at 250 variables). We then evaluated a \emph{random} baseline that uniformly-random samples as many parameter settings as there are Graph-Q-SAT validation evaluations and chooses the model with best validation performance. We find that this baseline had virtually identical performance to Graph-Q-SAT between 150 and 250 variables, making ti a possibility that the Graph-Q-SAT improvements at these sizes are occurring for similar reasons. We think the success of the random baseline is also independently interesting. It suggest that at least for this dataset, there are large areas in parameter space which contain good branching policies. This would be very surprising in vision for example. We would expect it would take an enormous number of random samples to, for example, build a better-than-random image classier. We hypothesize that GNN has a helpful inductive bias. Regardless of the particular parameter setting, similar variables will be mapped to similar values in a GNN. That means that similar variables will tend to be branched in succession since the highest scoring variables are likely to be similar to one another. This may be a good inductive bias as similar variables may be more likely to induce early conflicts if branched on in short succession. 

We note that the multiplicative improvement of Knuth Synthesis over \texttt{minisat} decreases with the number of variables. We suspect this is because the number of policy decisions is being held constant while the number of decisions needed to solve the problem is increasing exponentially.

\begin{table*}[t] 
\small
\centering
\setlength{\tabcolsep}{1.0pt}.
\begin{tabular}{llrrr}
\toprule
 Dataset & Size & \texttt{minisat} & \texttt{GraphQSAT} (1000s) & Ours (1000s) \\
\midrule
 \texttt{R3SAT} & 50 & 67 & 28 & \bf{21} \\
& 100  & 619 & 397 & \bf{134}\\
& 150 & 4144 & 3198 & \bf{1214} \\
& 200 & 26853 & 18178 & \bf{12794} \\
& 250 & 167993 & 114456 & \bf{95318} \\
\midrule
 \texttt{sgen} & 55 & 127374 & \bf{127943} & 182,241 \\
 \midrule
 \texttt{satfc-easy+satfc-hard} && 34410 & \bf{32626} &  41569 \\
\bottomrule
\end{tabular}
\caption{Improvement in tree size over \texttt{minisat} comparing Graph-Q-SAT to KnuthSynthesis.}
\label{tab:graphq_v_ks}

\end{table*}

We also ran a similar experiment evaluating on our \texttt{sgen-55} and \texttt{satfc-easy}+ \texttt{satfc-hard} datasets. For \texttt{satfc} there were a few especially difficult instances which caused extremely long training episodes in Graph-Q-SAT, which dominated the runtime. We removed those instances in our evaluation and reran our training runs. See Table \ref{tab:graphq_v_ks} for the results. In this case, Graph-Q-SAT showed moderate improvements while Knuth Synthesis struggled. We think this is very likely because the subsolver in this case (\texttt{minisat}) is a CDCL solver which breaks our tree MDP assumption (i.e., subproblems are independent). Knuth Synthesis prevents \texttt{minisat} from sharing clauses across subproblems and forces the VSIDs heuristic to cold start for every subproblem. In the case of \texttt{sgen}, we could not even improve upon \texttt{minisat} on a given instance during offline training.% relative to the minisat baseline so we had no hope of learning a policy. 
This phenomenon did not happen in the \texttt{R3SAT} dataset likely because CDCL performs very poorly in that setting and there is little value for sharing clauses across subproblems. We also note that the success of cube-and-conquer algorithms shows that is not always detrimental to solve subproblems independently with CDCL \cite{heule2011cube}.

\subsection{CDFs of complete results for tree size}

\begin{figure}
    \centering
    \includegraphics[width=\textwidth]{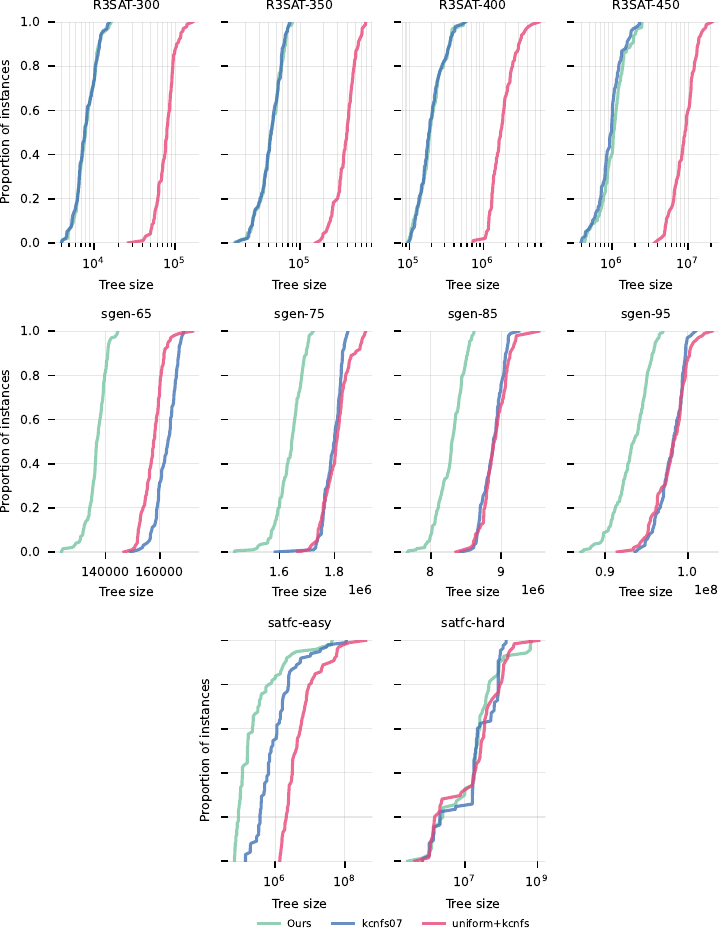}
    \caption{Tree size CDFs for the results in Table \ref{tab:results}. Comparison of Knuth Synthesis (Ours), \texttt{kcnfs07}, and \texttt{uni+kcnfs} on our 10 benchmarks.}
    \label{fig:cnfs}
\end{figure}

See Figure \ref{fig:cnfs}.

\subsection{Demonstrating \texttt{kcnfs07} is a strong baseline}

\texttt{kcnfs07} is a strong baseline for \texttt{R3SAT} and \texttt{sgen}, however CDCL solvers are much faster on \texttt{satfc}. It is well known that state-of-the-art industrial solvers (CDCL solvers with VSIDS-style branching heuristic) perform very poorly on  \texttt{R3SAT}.  \texttt{hkis} was the fastest solver on unsatisfiable instances at the 2021 SAT competition \citep{SATcomp2007} and \texttt{kcnfs07} was $>$100$\times$ faster on  \texttt{R3SAT}. Such strong industrial solvers also performed poorly on \texttt{sgen} (\texttt{kcnfs07} $>$3$\times$ faster). See Table \ref{tab:hkis} for complete results.

To better contextualize our results, we also evaluated the strong general-purpose DPLL heuristic Jeroslav-Wang (JW) \citep{jeroslow1990solving} for top-level decisions in the same way we benchmarked our method (i.e., \texttt{JW+kncfs}). \texttt{kcnfs} was substantially faster in most cases; see Table \ref{tab:jw} for complete results.
%\subsection{\texttt{hkis} v.s. \texttt{kcnfs07} on \texttt{sgen} and \texttt{R3SAT}}

\begin{table}[H]
\centering
\begin{tabular}{lrr@{\hskip 0.4cm}rr}
\toprule
& \multicolumn{2}{c}{\bf Tree size (1000s)} & \multicolumn{2}{c}{\bf CPU time (s)}\\
Distribution & \texttt{hkis}  & \texttt{kcnfs07} & \texttt{hkis}  & \texttt{kcnfs07}\\
\hline
\texttt{R3SAT-300} & 1,717.0 & 44.3 & 47.4 & 1.1\\
\texttt{R3SAT-350} & 15,120.8 & 215.7 & 836.4 & 6.1\\
\texttt{R3SAT-400} & - & 1,103.5 & > 6 hours & 30.8\\
\texttt{R3SAT-450} & - & 5,207.9 & > 6 hours & 168.9\\
\hline
\texttt{sgen-65} & 422.1 & 162.3& 12.2 & 4.0 \\
\texttt{sgen-75} & 2,996.7 &1,792.3 & 127.7 & 44.1 \\
\texttt{sgen-85} & 111,435.8 & 97,979.7 & 699.1 & 219.9\\
\texttt{sgen-95} & - & 97,979.7 & > 6 hours & 1,178.6 \\
\bottomrule
\end{tabular}
\caption{Mean tree size (1000s of nodes) and running time (CPU+GPU seconds) comparing \texttt{kcnfs} 
and \texttt{hKis}.}
\label{tab:hkis}
\end{table}

\begin{table}[H]
\centering
\begin{tabular}{lrrr}
\toprule
Distribution & \texttt{JW+kcnfs} Tree size (1000s) & \texttt{JW+kcnfs} CPU time (s) & Slower than \texttt{kcnfs}\\
\hline
\texttt{R3SAT-300} & 16.9 & 4.5 & $3.7\times$\\
\texttt{R3SAT-350} & 87.6 & 14.5 & $2.4\times$\\
\texttt{R3SAT-400} & 409.8  & 62.9 & $2.0\times$\\
\texttt{R3SAT-450} & 1,977.6 & 336.0 & $2.0\times$\\
\hline
\texttt{sgen-65} & 199.7 & 5.5 & $2.6\times$\\
\texttt{sgen-75} & 2,223.7 & 29.90 & $1.3\times$\\
\texttt{sgen-85} & 10,915.0 & 134.7 & $1.2\times$\\
\texttt{sgen-95} & 119,200.8 & 1433.3 & $1.1\times$\\
\bottomrule
\end{tabular}
\caption{Mean tree size (1000s of nodes) and running time (CPU+GPU seconds) for \texttt{JW+kcnfs}.}
\label{tab:jw}
\end{table}

\subsection{Challenging out-of-distribution generalization on \texttt{satfc}}

We evaluated generalization of our \texttt{satfc} model on more challenging \texttt{satfc} instances (\texttt{satfc-hard}). These were instances where \texttt{kcnfs} took more than 100,000 decisions (as opposed to our training set where instances were solvable in $<$100,000 decisions. Bottom of Figure \ref{fig:cnfs} illustrates how different the runtime distributions are between the easy and hard instances. Our model outperforms the \texttt{uni+kcnfs} baseline but could not  match the performance of \texttt{kcnfs}. These results are almost entirely driven by the performance on a few especially hard instances as the CDFs show in Figure \ref{fig:cnfs}. See Table \ref{tab:satfc_hard}.

\begin{table*}[t] 
\small
\centering
\setlength{\tabcolsep}{2.0pt}.
\begin{tabular}{lll@{\hskip 0.2cm}rrr@{\hskip 0.6cm}rrr}
\toprule
&  & & \multicolumn{2}{c}{\bf Tree size (1000s) [Reduction] } &  \multicolumn{3}{c}{\bf CPU+GPU time (s) [Reduction] } \\
Distribution & In/Out & \texttt{uni+kcnfs} & \texttt{kcnfs07} & Ours & \texttt{uni+kcnfs} & \texttt{kcnfs07} & Ours \\
\midrule
\texttt{satfc-hard} &  Out & 5,914.3 &  { \bf 3,201.8} & 3,538.7 &  259.2 & {\bf 147.7} & 166.3 \\
\bottomrule
\end{tabular}
\caption{Mean tree size (1000s of nodes) and running time (CPU+GPU seconds) for hard \texttt{satfc} instances.}
\label{tab:satfc_hard}
\end{table*}

\section{Benchmark Details}
\label{app:benchmarks}

For \texttt{R3SAT}, we followed \citet{Crawford1996} to estimate the location of the phase transition with a number of clauses-to-number of variables $n = 4.258 \cdot m + 58.26 \cdot m^{-2/3}$. We used the \texttt{CNFgen} \citep{Lauria2017} python package to generate.

\texttt{sgen} \citep{Spence2010} is a hand-crafted generator that is notoriously difficult for its size. An \texttt{sgen} generated instance was the smallest to be unsolved at the 2009 SAT Competition. The principle behind the generator is to ensure that many assignments must be made before reaching a conflict. At a high level, the variables are partitioned so that variables across partitions seldom appear in clauses together but contradictions occur across partitions so it is difficult to find a contradiction without assigning many variables. We set the \texttt{-unsat} option on the generator, which guarantees that generated instances that contain contradictions.

For \texttt{satfc} \citep{Frechette2016}, we used this publicly available simulator of the 2016 FCC spectrum auction: \url{https://github.com/newmanne/SATFC/tree/development/simulator}
. We ran 160 different simulator configurations and collected all of the station repacking instances. We solved each instance and filtered satisfiable instances. The FCC spectrum auction represented the airspace for the entirety of the USA. To create smaller problems that our architecture could handle, we built subsets of the USA station interference graph. For each subset graph, we started with all the stations from one of the following prominent american cities: NEW YORK, LOS ANGELES, CHICAGO, HOUSTON, PHILADELPHIA, PHOENIX, SAN ANTONIO, SAN DIEGO, DALLAS, SAN JOSE. We then added stations that were \texttt{\textcolor{red}{Num links}} hops away from our initial set of stations in the full interference graph. For each starting city, we created interference graphs for each setting of \texttt{\textcolor{red}{Num links}} from 1 to 4, leaving us with 40 interference graphs in total. For each graph, we ran four separate experiments with standard deviation of noise of the station valuation model to \{0.01, 0.05, 0.1, 0.5\}. Each auction simulation took at most one hour to complete.

For a given \texttt{\textcolor{red}{City}}, \texttt{\textcolor{red}{Num links}}, and \texttt{\textcolor{red}{Noise}}, we ran the following call string: \\
%\begin{lstlisting}
\texttt{build/install/FCCSimulator/bin/FCCSimulator -CONFIG-FILE \\SATFC/satfc/src/dist/bundles/ satfc\_8.yaml -MAX-CHANNEL 29\\ -MAX-CHANNEL-FINAL 36  \\-VOLUMES-FILE  src/dist/simulator\_data/volumes.csv -POP-VALUES true -START-CITY [\textcolor{red}{City}] -CITY-LINKS [\textcolor{red}{Num links}] -CNF-DIR ./cnfs \\-UHF-ONLY true -INCLUDE-VHF false -NOISE-STD [\textcolor{red}{Noise}]}
%\end{lstlisting}

We solved each unsatisfiable instance with \texttt{minisat} \citep{Een2003} and recorded the number of decisions needed to solve each instance. \texttt{satfc} was each instance that took less than 100,000 decisions. There were some instances that \texttt{kcnfs07} immediately crashed on for unknown reasons for which we removed.

\end{document}